# Information Stealing in Federated Learning Systems Based on Generative Adversarial Networks

Yuwei Sun
*Graduate School of Information Science and Technology*
*The University of Tokyo*
Tokyo, Japan
sywtokyo@hongo.wide.ad.jp

Ng S. T. Chong
*Campus Computing Centre*
*United Nations University*
Tokyo, Japan
ngstc@unu.edu

Hideya Ochiai
*Graduate School of Information Science and Technology*
*The University of Tokyo*
Tokyo, Japan
jo2lxq@hongo.wide.ad.jp

*Abstract*—An attack on deep learning systems where intelligent machines collaborate to solve problems could cause a node in the network to make a mistake on a critical judgment. At the same time, the security and privacy concerns of AI have galvanized the attention of experts from multiple disciplines. In this research, we successfully mounted adversarial attacks on a federated learning (FL) environment using three different datasets. The attacks leveraged generative adversarial networks (GANs) to affect the learning process and strive to reconstruct the private data of users by learning hidden features from shared local model parameters. The attack was target-oriented drawing data with distinct class distribution from the CIFAR-10, MNIST, and Fashion-MNIST respectively. Moreover, by measuring the Euclidean distance between the real data and the reconstructed adversarial samples, we evaluated the performance of the adversary in the learning processes in various scenarios. At last, we successfully reconstructed the real data of the victim from the shared global model parameters with all the applied datasets.

## I. Introduction

Privacy is a dominant concern in different aspects of the digital world, so is the case in deep learning (DL) systems. Though the current digitalization and knowledge acquisition happening in various industries is bringing us to a new era of smart society, the two central questions have been how to protect personal data while enabling compliance with training model sharing regulations.

Federated learning (FL) has been used as a collaborative learning scheme for users with partial and imbalanced data to train a DL model for a shared goal. The privacy-oriented nature of FL allows users to collaborate through sharing a local model, as a result, without disclosing their private data. Unfortunately, even within such a decentralized learning scheme, an adversarial attack launched by an endpoint attacker could effectively exacerbate the privacy and security of it, revealing the private data of users.

An adversarial attack on FL involves several aspects such as manipulating endpoint training data and falsifying a local training model. We focused on the latter to affect the learning process and constructed target-oriented adversarial attacks based on generative adversarial networks (GANs). The compromised local model of the adversary including malicious model parameters was shared with the victim through the parameter server, then the local model of the victim would need more effort in training on the misguided contents to counteract the defect, which exposed more details to the adversary (Fig. 1). In our settings, the adversary had access to the latest global model and reached consensus on the learning goal with the other users. The contributions of this research include:

- Each user was assigned with a distinct data class for the collaborative learning, instead of randomly selected data from all classes, which added more difficulty to mounting an attack.

- We optimized the structure of GANs, modifying the balance between the learning rate of the discriminator and the one of the generator.

- We mounted attacks with three different datasets on a two-user scenario and an 11-user scenario respectively. Different from the former research [1], we studied the impact of such attacks on the systems' regular classification tasks against the accuracy, precision, recall, F1 Score, and ROC curve, measuring the performance of the adversary in data reconstruction by introducing Euclidean distance.

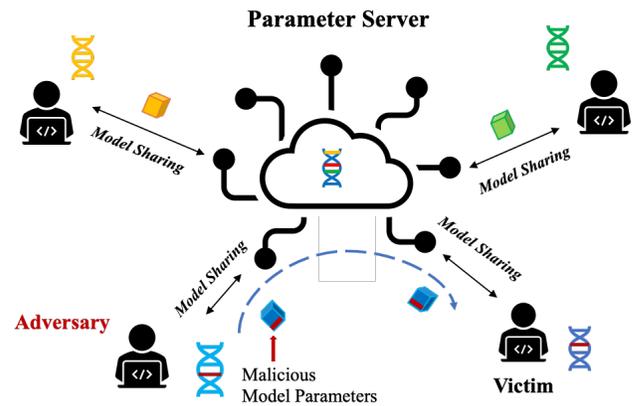

Fig. 1. Information stealing in federated learning through malicious model parameters sharing.

This paper is organized as follows. Section 2 discusses related work on adversarial attacks on DL and FL systems.

Section 3 presents the mechanism of the GANs-based threat model in FL. Section 4 presents the performance evaluation of the attacking scheme based on a range of metrics with three datasets of CIFAR-10 [2], MNIST [3] and Fashion-MNIST [4]. Section 5 discusses the pros and cons of the threat model. Finally, in section 6, we conclude the paper and give out the future work of this study.

## II. RELATED WORK

Despite the extensive applications and utilization of DL, many researchers have shown an adversary can manipulate the decision result of the trained model, and even steal original data from the model. Unlike centralized learning schemes, FL is a distributed learning approach that allows multiple actors to collaboratively build a global model without disclosing private data. Unfortunately, recent research has shown that FL systems are often vulnerable to adversarial attacks such as data poisoning, backdoor attacks, etc. An adversary can perturb an FL system with sufficient shared gradients [1][5]. For instance, Zhang et al. [6] proposed a backdoor poisoning attack on FL to produce a global model that performs high accuracy on its main task while also exhibiting good attacking performance on targeted inputs. Cao et al. [7] presented a distributed poisoning attack on FL based on label-flipping, with 10 users involved, and researched two key factors affecting the attack success rate, the numbers of poisoned samples and attackers.

On the other hand, the implementation of GANs has been increasingly used to defeat machine learning algorithms [8][9]. For instance, Fredrikson et al. [10] designed a threat model against a facial recognition system, reconstructing the facial images of a participant through accessing the application programming interface (API). Additionally, Zhang et al. [11] presented an adversarial approach to violate the health data of patients in cloud-assisted e-health systems.

Whereas former research has focused on the success rate of an attack, our efforts focus on measuring reconstruction quality. We conducted target-oriented adversarial attacks based on GANs using two datasets of MNIST and Fashion-MNIST, with different difficulties for perturbing the system, to evaluate the efficiency of the threat model. Additionally, we studied how the attack would affect the performance of the other users and the global model, which is also different from former research, where weights were put on the adversary's performance.

## III. ADVERSARIAL ATTACKS ON FEDERATED LEARNING WITH GENERATIVE ADVERSARIAL NETWORKS

### A. Dataset

We employed three datasets for conducting the experiments, namely, CIFAR-10, MNIST and Fashion-MNIST. These datasets pose different degrees of difficulty for perturbing the FL system. CIFAR-10 [2] is a collection of 10 types of objects' color images, covering 50,000 training samples labeled as airplane, automobile, and so on and 10,000 test samples. The size of the images in it is 32×32. We converted all the color images into grayscale images with a size of 28×28 for simplicity. MNIST [3] is a handwritten digit image dataset containing 50,000 training samples labeled as 0-9, and 10,000 test samples. The size of the images in it is 28×28. Fashion-MNIST [4] is an image collection of 10 types of clothing containing 50,000 training samples labeled as shoes, t-shirts, dresses, and so on and 10,000 test samples. The size of the images in it is 28×28. Each data class of MNIST and Fashion-MNIST includes 5,000 training samples and 1,000 test samples.

### B. Federated Learning

Federated learning (FL) is a collaborative learning scheme that shares intelligence on local model training across the participating users to generate a global model, while the users' private data never leave the nodes. In detail (Fig. 2), first, users download the latest global model from the central server, thus updating their local models. Secondly, a subset of the participants is selected to conduct the model training using local datasets. The trained models are then uploaded to the parameter server for aggregation, where the global model is stored and maintained. There are several methods used for the aggregation, and the commonest one is the averaging aggregating, i.e., computing the average values of all the uploaded local parameters.

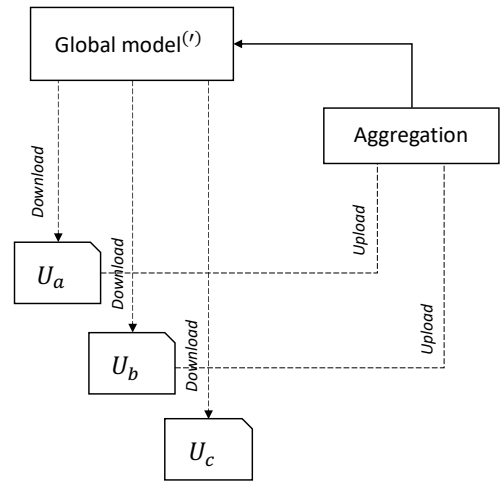

Fig. 2. Mechanism of federated learning for local model parameter sharing.

In our study, we considered a scenario where several participants worked together to solve a multi-cluster classification problem. Each participant only had access to limited data from a single cluster. For instance, with MNIST, each participant had the images of a specific digit and the common goal of the users was to train a global model with the ability to identify all the ten image classes. By sharing trained local model parameters, they achieved a global model with satisfying performance in the multi-cluster classification task. Additionally, participants agreed on the same training model architecture and hyper-parameters such as the learning rate, the number of update epochs at each iteration, and so on.

### C. Threat Model: Adversarial Attacks with GANs

The generative adversarial networks (GANs) involve both a generator network and a discriminator network. GANs are usually employed to generate highly imitated images and data distributions through a race between the discriminator and the generator. The discriminator is trained to distinguish adversarial samples generated by the generator from real data of a user (Fig. 3). Each party learns to improve its performance as much as possible by minimizing the prediction loss. The loss of the generator is computed from the prediction results of the generated samples and real data. And the loss of the discriminator is computed from the prediction results and the ground truths. The tradeoff for updating the model could be represented with Equation (1).

$$\min_G \max_D V(D, G) = \mathbb{E}_{x \in p_{target}(x)}[\log D(x)] + \mathbb{E}_{z \in p_{noise}(z)}[\log(1 - D(G(z)))] \quad (1)$$

Where $G$ represents the generator, $D$ represents the discriminator, $x$ is the original input data, $z$ is gaussian noise used as the input of the generator, and $G(z)$ is the generated adversarial samples.

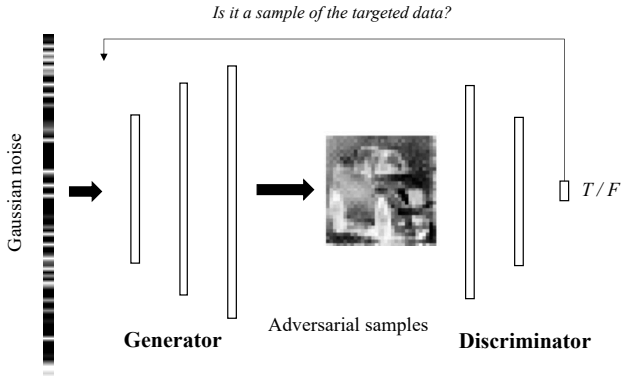

Fig. 3. The tradeoff between the discriminator and the generator: the discriminator is trained to distinguish the adversarial samples drawn from the generator from targeted real data of the victim.

By employing GANs, an adversary in a collaborative learning scheme such as FL could steal information on local training data from the other users by just downloading the latest global model, which includes the aggregated parameter information of the victim's local training model. To implement GANs in an adversary's training scheme, first, the adversary replaces the discriminator of implemented GANs with the latest global model from the parameter server. Then the generator generates adversarial samples from Gaussian noise and updates itself based on the computed loss between the inference result of the adversarial samples from the discriminator and the targeted classes, as defined in (2).

$$L_G(\theta_g) = \mathbb{E}_{z \in p_{noise}(z)}[\log(D(G(z)))] \quad (2)$$

Where $L_G$ is the loss of the generator, $\theta_g$ is the model of the generator, z is the input gaussian noise of the generator, and G(z) is the generated adversarial sample.

We employed a four-layer deep convolutional neural network as the architecture for both the local training model and the discriminator of the GANs to conduct the federated learning and further mount the attack (Fig. 4). The first convolutional layer has a convolution kernel of size 3×3 with a stride of 2 and it takes one input plane and it produces 32 output planes, followed by a ReLU activation function. Whereas the second convolutional layer takes 32 input planes and produces 64 output planes and it has a convolution kernel of size 3×3 with a stride of 2, followed by ReLU. Then the output is flattened to a size of, where a linear transformation is applied which takes as input the tensor and outputs a tensor of size 200. A ReLU activation function is applied to the output, which is then followed by another linear transformation which takes as input the tensor of size 200 and outputs a tensor of size 11, where the 11th output is associated with the generated results by the adversary.

Another four-layer deep convolutional neural network was employed as the architecture of the generator of GANs (Fig. 4). Generator takes as input a 100-dimensional uniform distribution followed by a linear transformation with an output of size 1568, and a ReLU activation function. Then the output is reshaped to 32 planes with a size of 7×7. A transposed convolution kernel of size 3×3 with a stride of 2 is applied, followed by ReLU. Then another transposed convolution kernel of size 3×3 with a stride of 2 is applied, taking 32 input planes and producing 16 output planes, followed by ReLU. Finally, a transposed convolution kernel of size 3×3 with a stride of 1 followed by a Tanh activation function converts the input to a 28x28 image as output.

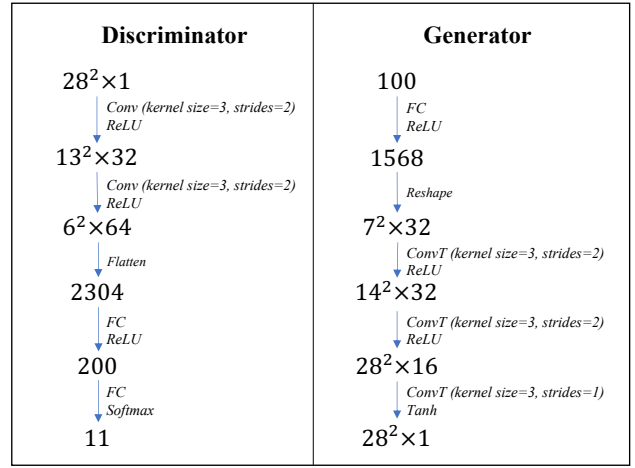

Fig. 4. The architecture of the discriminator and the generator.

Furthermore, we employed the Adam learning function to update the model parameters as defined in (3). Through experiments, it showed that applying a relatively larger learning rate of the discriminator with respect to the learning rate of the generator yielded a better attack performance. For this reason, we optimized the learning rates and employed values of 0.005 and 0.001 for the discriminator and generator, respectively. We employed a batch size of 16 and an epoch of one to update the local model at each round, where a small batch size contributes to a crispier reconstruction result.

$$m_t = \beta_1 * m_{t-1} + (1 - \beta_1) * \frac{\partial L}{\partial W_t}$$

$$V_t = \beta_2 * v_{t-1} + (1 - \beta_2) * \frac{\partial L}{\partial W_t} \odot \frac{\partial L}{\partial W_t}$$

$$\widehat{m_t} = \frac{m_t}{1-\beta_1^t} \qquad (3)$$

$$\widehat{v_t} = \frac{v_t}{1-\beta_2^t}$$

$$W_{t+1} = W_t - \eta * \frac{1}{\sqrt{\widehat{v_t}}+\epsilon} \odot \widehat{m_t}$$

Where $L$ is the training loss, $W$ is weights of the node, $\eta$ is the learning rate, $\beta_1$ and $\beta_2$ are the exponential decay rates with values of 0.5 and 0.999 respectively, and $\epsilon$ is used to prevent a zero-valued denominator, with a value of 1e-7.

To successfully reconstruct the victim's local data from model parameters, we mounted the attack upon the systems reaching a validation accuracy better than 0.90 (Fig. 5). For each round, we employed the test set from the applied dataset for the validation. Besides, for the adversarial training of the GANs at each round, the adversary conducted 500 epochs of update with a batch size of 16. Then we generated adversarial samples based on the trained generator, and used these samples as the input to train the adversary's local model with a label of the 11th class.

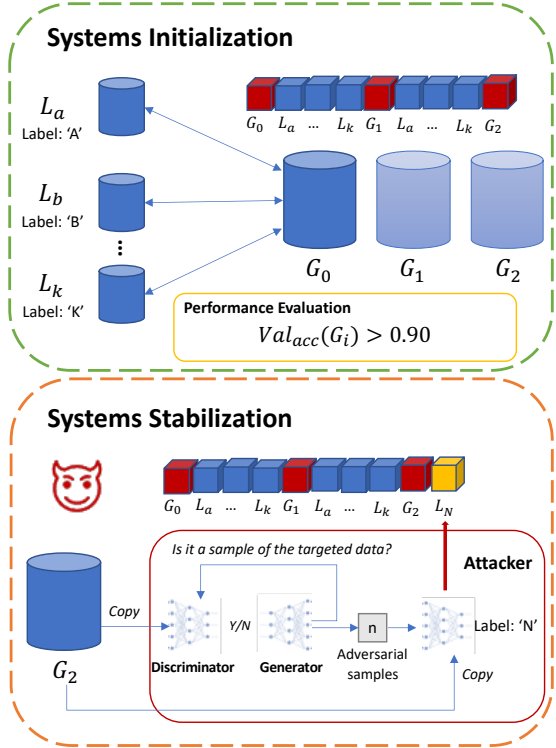

Fig. 5. Threat model: the adversary mounted the target-oriented adversarial attacks on FL upon the systems reaching a validation accuracy better than 0.90.

The adversary updates the model of its discriminator network by downloading the latest global model from the parameter server, as a result, generating crispier samples to train its local model with misguiding labels. The injection of the malicious model parameters is aggregated to the global model and downloaded by the victim. Consequently, the compromised global model would lure the victim to expose more detail on the local data due to the victim needing more effort on identifying the adversarial data from the real local data. On the other hand, the generator performed better and better, finally, generated a crispy image that revealed the victim's local training data.

## IV. EVALUATION

### A. Experiment Setup

First, we considered a simple scenario with only two users involved, the victim and the adversary, namely a two-user scenario. Then, a more complicated scenario, an 11-user scenario, was employed. In this scenario, a total of 10 users were conducting model training based on distinct sets of 5,000 images from one data class in the training set. In contrast, the adversary didn't possess training data from the dataset, instead, it trained on a total of 5,000 adversarial samples generated by the generator at each round. Besides, for each round, a user randomly selected 50 images as the input to conduct local model training. We applied the CIFAR-10 to the two-user scenario, and MNIST and Fashion-MNIST to the 11-user scenario. For each attack attempt, we assigned the adversary a specified target, i.e., a specific data class of the victim.

### B. Numerical Results

The attack was initialized after the performance of the global model reached 0.90. To evaluate the systems' performance at initialization and stabilization respectively, we employed the ROC curve (receiver operating characteristic curve), a graph showing the performance of a classification model through the relationship between the true positive rate (TPR) and the false positive rate (FPR), as defined in (4). Moreover, AUC (Area under the ROC Curve) measures the entire area underneath the ROC curve, providing a measurement approach to model performance across all possible classification thresholds, where a higher AUC score represents a better performance result. The ROC curves and computed AUCs with respect to the global model's performance for each data class when applying the MNIST on the 11-user scenario are shown below, which were evaluated upon the systems initializing and reaching the desired performance of 0.90 (Fig. 6).

$$TPR = \frac{TP}{TP+FN} \qquad (4)$$

$$FPR = \frac{FP}{FP+TN}$$

Where $TP$ (True Positives) and $FP$ (False Positives) indicate the number of relevant images correctly and incorrectly classified by the model, respectively. *TN* (True Negatives) and *FN* (False Negatives) indicate the number of irrelevant images correctly and incorrectly classified, respectively.

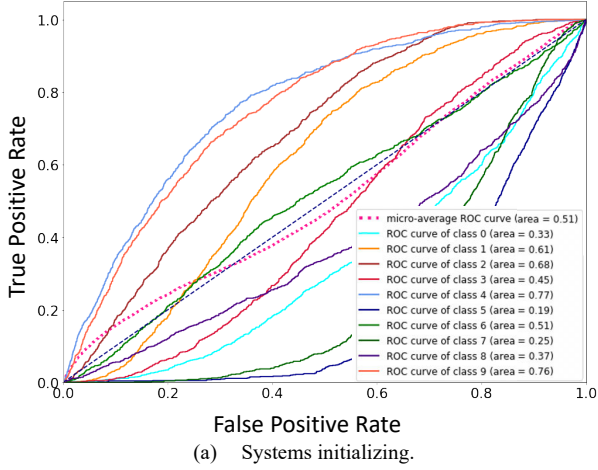

(a) Systems initializing.

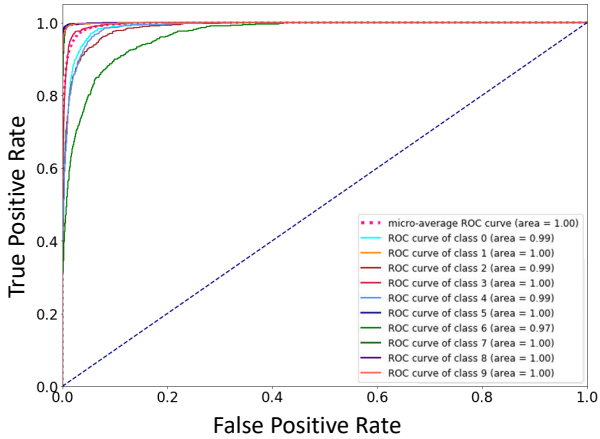

(b) Systems reaching the desired performance of 0.90.

Fig. 6. The ROC curves and AUCs with respect to the global model's performance for each data class with the MNIST on the 11-user scenario.

To evaluate the system's performance before and after mounting the adversarial attacks, we employed a range of metrics covering Accuracy, Precision, Recall, and F1-Score defined in (5). Precision represents the fraction of relevant data successfully classified among validation data; Recall represents the fraction of relevant images successfully classified among all relevant datas. F1-Score shows overall performance based on Precision and Recall. All metrics are macro values computed by each class and taken the average of all classes.

$$Precision = \frac{TP}{TP+FP}$$

$$Recall = \frac{TP}{TP+FN} \quad (5)$$

$$F1\text{-}Score = \frac{2 \times Precision \times Recall}{Precision + Recall}$$

We mounted the adversarial attacks with the MNIST on the 11-user scenario for a total of 100 rounds. We evaluated the performance of the global model at each round based on the aforementioned metrics, using the test set from the applied dataset (Fig. 7). As shown in the graph, after initializing the attack, the validation accuracy kept stable during the process while the precision, recall, and F1-Score showed a decrease due to the malicious parameters trained on the adversarial samples.

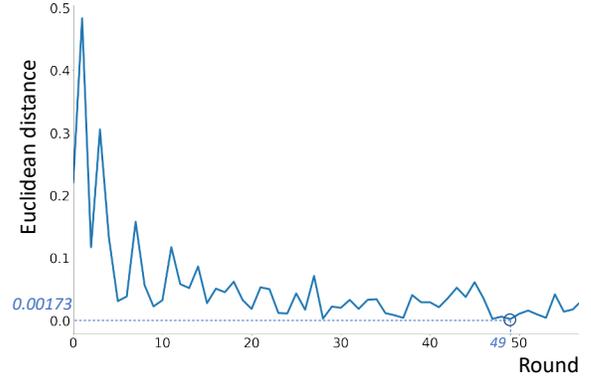

Fig. 7. Validation results of the global model at each round before and after mounting the attack at Round 42 (with the MNIST on the 11-user scenario targeting at the data class of 'digit one').

Furthermore, to evaluate the performance of the adversary for reconstructing real data at each round, we employed as a metric the Euclidean distance defined in (6). For each round's adversarial training, we computed the average Euclidean distance between the generated adversarial samples and the targeted real data of the victim (Fig. 8). As shown in the graph, the distance score kept decreasing, and after 49 rounds of the adversarial training, it reached the minimum score of 0.00173, significantly lower than the initial score of 0.221.

$$Dis(I_{fake}, I_{real}) = \sqrt{\sum(I_{fake} - I_{real})^2} \quad (6)$$

Where $I_{fake}$ represents adversarial samples, $I_{real}$ represents the targeted real data of the victim.

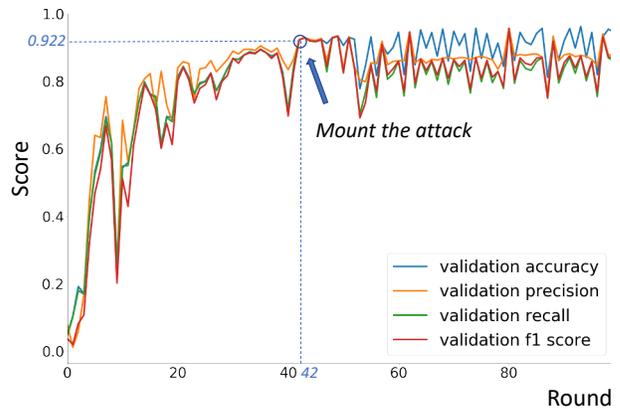

Fig. 8. The average Euclidean distances between the adversarial samples and the targeted real data of the victim at each round of training (with the MNIST on the 11-user scenario targeting at the data class of 'digit one').

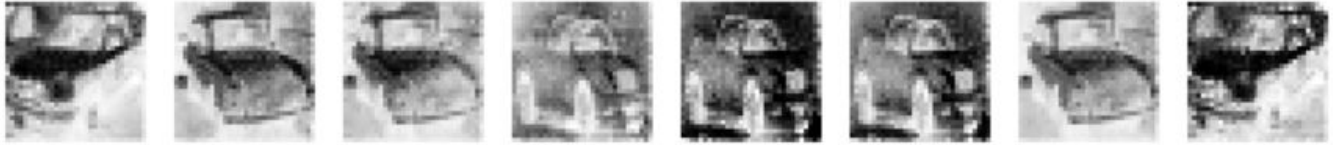

(a) Reconstruction results with the CIFAR-10 on the two-user scenario.

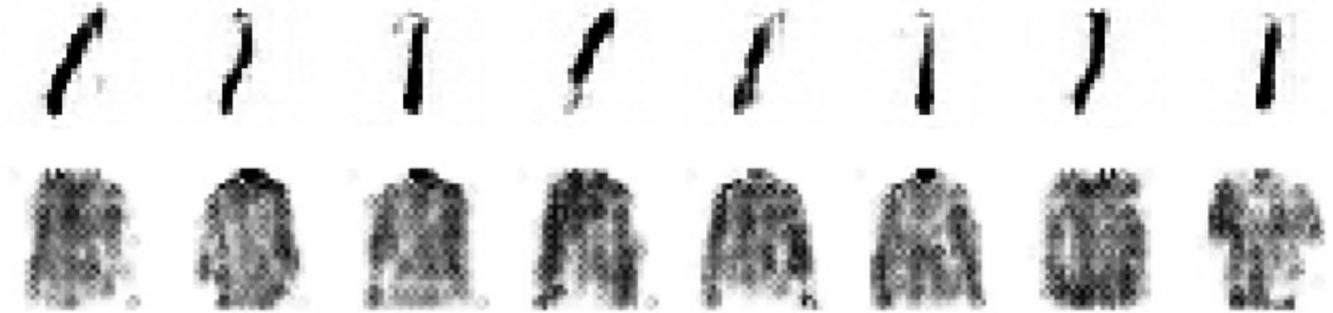

(b) Reconstruction results with the MNIST and the Fashion-MNIST on the 11-user scenario. The results are noisier compared with the results of the two-user scenario, however, still revealing sufficient details to identify the targeted real data.

Fig. 11. Reconstruction results of the GANs-based adversarial attacks on the two-user scenario and the 11-user scenario.

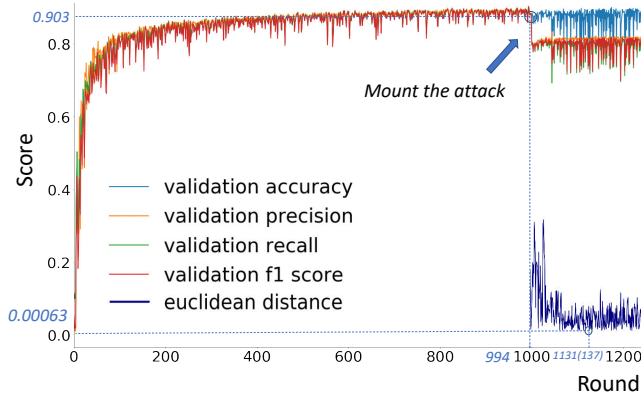

Fig. 9. Evaluation results with the Fashion-MNIST on the 11-user scenario for a total of 1250 rounds (targeting at the data class of 'shirt').

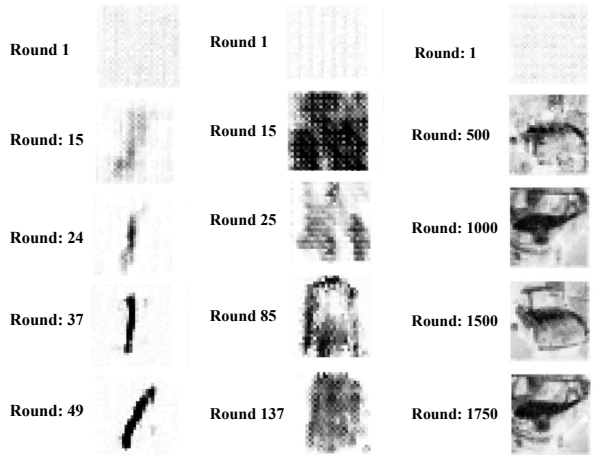

Fig. 10. Reconstruction processes of the adversarial samples based on the MNIST, Fashion-MNIST, and CIFAR-10 respectively.

Besides, we mounted the adversarial attacks with the Fashion-MNIST on the 11-user scenario for a total of 1250 rounds. The corresponding evaluation results based on the accuracy, precision, recall, F1-Score, and Euclidean distance are shown below (Fig. 9). The reconstruction processes and the final results of the adversarial samples are shown in Fig. 10 and Fig. 11, respectively.

## V. CONCLUSION

In this research, we successfully mounted adversarial attacks on various federated learning (FL) environments using the three different datasets, CIFAR-10, MNIST and Fashion-MNIST, respectively. Instead of randomly selecting data from the dataset as the local training data, we assigned data from a distinct class to a user for the learning, which added more difficulty to mounting the adversarial attack. Besides, we evaluated the global model's performance before and after mounting the attack based on a range of metrics, and the performance of the adversary by introducing the Euclidean distance score as a measurement of the reconstruction results. In the future, defense strategies for detecting and combating such attacks based on methods such as the blockchain, would be studied.